\documentclass[journal]{IEEEtran}

\ifCLASSINFOpdf
\usepackage[pdftex]{graphicx}
\else
\usepackage[dvips]{graphicx}
\fi
\usepackage[T1]{fontenc}
\usepackage{aecompl}
\usepackage{cite}
\usepackage[latin1]{inputenc}
\usepackage{colortbl}
\usepackage[cmex10]{amsmath}
\usepackage{amssymb}  
\usepackage{mathtools}
\usepackage{makecell}
\usepackage{booktabs}
\usepackage{multirow}
\usepackage{arydshln}
\usepackage{color, soul}
\usepackage[hidelinks]{hyperref}
\usepackage{enumerate}
\usepackage{subfigure}
\usepackage{threeparttable}
\usepackage{tabularx}
\usepackage{tabu}
\usepackage{algorithm}
\usepackage{algorithmicx}
\usepackage{algpseudocode}
\usepackage{soul}
\soulregister\cite7
\soulregister\ref7
\usepackage[dvipsnames]{xcolor}

\graphicspath{{Figures/}}
\usepackage{verbatim}
\begin{document}
	\title{HybrUR: A Hybrid Physical-Neural Solution for Unsupervised Underwater Image Restoration}
	\author{Shuaizheng~Yan, Xingyu~Chen, Zhengxing~Wu,~\IEEEmembership{Senior Member,~IEEE} Min~Tan, and Junzhi~Yu,~\IEEEmembership{Fellow,~IEEE}
		\thanks{This work was supported in part by the National Key Research and
			Development Program of China under Grant 2019YFB1310300; in part by
			Youth Innovation Promotion Association CAS (2019138) \emph{(Corresponding author: Zhengxing Wu.)}} 
		\thanks{S. Yan is with the Laboratory of Cognitive and Decision Intelligence for Complex System, Institute of Automation, Chinese Academy of Sciences, Beijing 100190, China, and also with the Department of Mechanical Engineering, Fuzhou University, Fuzhou 350000, China (e-mail: yanshuaizheng2018@ia.ac.cn)}
		\thanks{Z. Wu, and M. Tan are with the Laboratory of Cognitive and Decision Intelligence for Complex System, Institute of Automation, Chinese Academy of Sciences, Beijing 100190, China, and also with the School of Artificial Intelligence, University of Chinese Academy of Sciences, Beijing 100049, China (e-mail: zhengxing.wu@ia.ac.cn; min.tan@ia.ac.cn).}
		\thanks{X. Chen is with Xiaobing.AI, Beijing 100080, China (email: chenxingyu@xiaobing.ai).}
		\thanks{J. Yu is with the Laboratory of Cognitive and Decision Intelligence for Complex System, Institute of Automation, Chinese Academy of Sciences, Beijing 100190, China, and also with the State Key Laboratory for Turbulence and Complex Systems, Department of Advanced Manufacturing and Robotics, College of Engineering, Peking University, Beijing 100871, China (e-mail: junzhi.yu@ia.ac.cn).
	}}
	\maketitle
	
	\begin{abstract}
		Robust vision restoration of underwater images remains a challenge. Owing to the lack of well-matched underwater and in-air images, unsupervised methods based on the cyclic generative adversarial framework have been widely investigated in recent years.
		However, when using an end-to-end unsupervised approach with only unpaired image data, mode collapse could occur, and the color correction of the restored images is usually poor. In this paper, we propose a data- and physics-driven unsupervised architecture to perform underwater image restoration from unpaired underwater and in-air images. For effective color correction and quality enhancement, an underwater image degeneration model must be explicitly constructed based on the optically unambiguous physics law. Thus, we employ the Jaffe-McGlamery degeneration theory to design a generator and use neural networks to model the process of underwater visual degeneration. Furthermore, we impose physical constraints on the scene depth and degeneration factors for backscattering estimation to avoid the vanishing gradient problem during the training of the hybrid physical-neural model. Experimental results show that the proposed method can be used to perform high-quality restoration of unconstrained underwater images without supervision.	On multiple benchmarks, the proposed method outperforms several state-of-the-art supervised and unsupervised approaches. We demonstrate that our method yields encouraging results in real-world applications.
		
	\end{abstract}
	

	\section{Introduction}
	\begin{figure}
		\centering
		\includegraphics[width=1\linewidth]{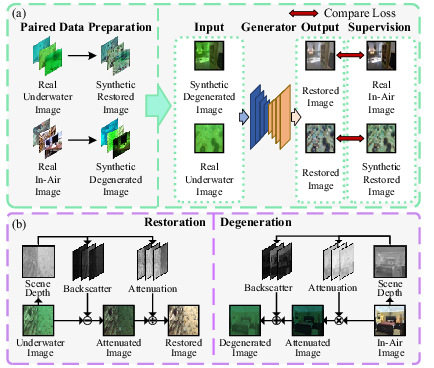} 
		\caption{Underwater image restoration in supervised and unsupervised training frameworks. (a) In most two-stage supervised schemes, synthetic images should be generated first to construct a paired dataset. (b) We develop an unsupervised framework, which is a combination of the Jaffe-McGlamery model and CycleGAN, for training unpaired underwater and in-air images in a single stage. Our model reconstructs the scene depth, attenuation, and scattering from a single image to achieve excellent image quality restoration.}
		\label{fig:1}
	\end{figure}
	\IEEEPARstart{W}{ith} the rapid development of underwater robots in recent years, exploration of the ocean has rapidly increased. Real-time and accurate perception of underwater scenes is vital in most underwater applications. Among numerous methods, the vision-based method is regarded as an efficient and informative approach \cite{Foresti2000, Myint2018}. However, the quality of underwater images is poor because of optical degeneration, including light scattering and absorption, which lead to difficulties in underwater visual perception \cite{ChenTCSVT2021,ORBSLAM}.
	
	To tackle these challenges, underwater image enhancement and restoration methods based on convolutional neural network (CNN) have been widely studied, where good results have been achieved by some supervised approaches \cite{Islam2020, Chen2021}. As shown in Fig.~\ref{fig:1}(a), the supervised approaches are usually with two-stage training for the lack of ground truth, i.e., synthesizing pseudo ground truth and training an end-to-end generator. In the first stage, traditional image processing methods \cite{DCP2010, WCID2011, UDCP2013, Ancuti2017, Peng2017} or deep learning algorithm \cite{Wang2017ICIP, LiCY2020, Chen2019} are employed to synthesize pseudo in-air images from real underwater images. Conversely, synthetic underwater images can also be generated from real in-air images \cite{Wang2019,WaterGAN}. In this manner, matched pairs of real images and pseudo ground truth are generated for end-to-end supervised learning. Intuitively, the training of supervised methods is costly in data preparation. Moreover, low-quality synthetic samples could have an adverse impact on the model performance.
	
	Recently, researchers have attempted to employ style transfer frameworks such as cyclic generative adversarial network (CycleGAN) for underwater visual enhancement, where unpaired underwater and in-air data can be directly utilized for learning \cite{LiCY2018, Chen2016, Fabbri2018}. However, few unsupervised method has achieved impressive results \cite{LiCY2018}, which is usually poor in detail generation and color correction. In addition, for the training of the unsupervised framework, the problem of mode collapse usually harms the learning stability, leading to distorted content and odd patches. In general, the mode collapse is produced by the lack of a specific architecture or constraint in the end-to-end unsupervised framework to preserve the image content, allowing the generator to break the image structure \cite{Zhu2017, Lee2018}.
	
	In this paper, we argue that the degeneration process of most underwater images can be expressed in explicit physical models according to different optical degeneration theories. In previous literature, optical models have been applied to produce pseudo in-air image samples, but few of them have been combined with unsupervised training architectures \cite{Wang2019}. Referring to the Jaffe-McGlamery model, image degeneration is caused by light absorption and scattering, both of which are exponentially related to the scene depth. Therefore, we assume that we can derive the expression of the image color correction process from the Jaffe-McGlamery degeneration model and construct a cyclic generative framework that follows explicit physics-based formulas. As shown in Fig.~{\ref{fig:1}(b)}, attenuation and backscatter maps that are highly associated with the scene depth can be disentangled from an underwater image to facilitate image restoration. According to representation disentanglement theory \cite{Chen2016}, we develop a closed training loop based on image degeneration and restoration processes, and divide a single generator into four sub-networks to estimate attenuation, scattering, veiling light (i.e., background illumination), and depth map. Instead of separated learning processes, all sub-networks are integrated in a cyclic architecture with the forward-inverse Jaffe-McGlamery model and trained in an end-to-end manner. Specifically, the scene depth sub-network is designed with a typical encoder-decoder network, and other three encoder-based sub-networks are employed to learn the attenuation, scattering, and veiling light in RGB format.
	
	To the best of our knowledge, this paper is the first attempt to jointly learn a physical-neural underwater image restoration model using an unsupervised framework. Additionally, to address the problem of mode collapse during training, we propose a novel physics-based learning paradigm by integrating underwater optical properties into the training process. First, we introduce a concatenated input with the original image and an estimated depth map for the attenuation and scattering sub-networks, where the correlation between the depth and degeneration factors can be derived for stable training. Second, a backscatter loss function that relies on the backscatter estimation is presented to improve the joint optimization of sub-networks. We validate the effectiveness of the proposed framework and physics-based learning paradigm through qualitative and quantitative comparison experiments.
	
	The contributions of this study are summarized as follows:
	\begin{itemize}
		\item[$\bullet$]
		We propose a hybrid physical-neural framework for underwater image restoration that can be trained in an unsupervised manner without the need of paired underwater and in-air images.
		\item[$\bullet$]
		We develop a physics-based learning paradigm by integrating the network with underwater optical properties that can avoid training collapse and improve the performance of visual restoration.
		\item[$\bullet$]
		Extensive experiments based on the RUIE, U45, and UIEB datasets show that the proposed method outperforms the state-of-the-art underwater image restoration or enhancement methods in terms of various metrics and qualitatively visualization. 
	\end{itemize}
	
	The remainder of this paper is organized as follows. In Section~II, related works are briefly introduced. The theoretical model and methodology are presented in Section~III. In Section~IV, a particular ablation study and the comparative experimental results of our method and state-of-the-art approaches are presented. Finally, the conclusions and future work are summarized in Section~V.

	\section{Related Works}
	To assess our contributions in relation to the vast literature on underwater image restoration and quality enhancement in the field of deep learning, we consider two aspects for each approach: whether the pseudo ground truth is utilized in training and whether the method refers to a certain physical model. Previous research can be broadly divided into three categories based on the availability of models and supervision. 
	
	\subsection{Supervised Enhancement}
	Before the emergence of deep learning-based methods, researchers mainly studied pixel-level operations to improve underwater image quality. These methods, including classical histogram equalization, white balance, wavelet transform, etc.\cite{Ancuti2017, Ancuti2012}, mainly refer to the numerical equalization of pixels and do not relate to specific scene depths or illumination, resulting in serious artifacts. Nevertheless, these traditional methods make it possible to train an end-to-end deep neural network in a supervised manner. Islam \emph{et al.} \cite{Islam2020} presented a conditional generative adversarial network-based model for real-time underwater image enhancement, and formulated multi-objective loss functions that evaluated the perceptual image quality in supervised training. Wang \emph{et al.} \cite{Wang2017ICIP} proposed a dehaze-based underwater image enhancement network (UIE-Net), which can effectively remove the haze effect caused by backscattering. Wang \emph{et al.} \cite{Wang2019} proposed an underwater generative adversarial network (UWGAN-UIE) to address the issues of color correction and dehazing using two individual modular networks. Hou \emph{et al.} \cite{Hou2018} combined a CNN with a gray-world algorithm to refine image blurring due to backward scattering. Li \emph{et al.} \cite{LiCY2020} proposed an underwater image enhancement CNN model based on prior underwater scenes to replace the traditional image processing method. 
	
	\subsection{Supervised Restoration}
	In contrast, model-based methods usually have strong interpretability through a certain physical model; however, their restoration results vary significantly among different models. In general, optical models that are widely employed in image restoration include the modified atmospheric turbulence, dehaze, and Jaffe-McGlamery degeneration models \cite{Chen2019, Chen2021, Akkaynak2019}. Most traditional model-based approaches are poor in color correction for extreme aquatic environments \cite{TPAMI2020}. To solve this problem, Akkaynak \emph{et al.} proposed Sea-Thru based on a revised Jaffe-McGlamery model and obtained nearly realistic restored images using an underwater depth map \cite{Akkaynak2019}. However, the Sea-Thru algorithm relies heavily on the accuracy of depth estimation; thus, unacceptable color distributions are produced when combined with off-the-shelf depth estimation networks. In addition, most model-based restoration approaches are limited by parameter search or optimization, which requires a large amount of computation costs.
	
	Learning approaches with an optical model have been investigated in recent years. WaterGAN and UWGAN \cite{WaterGAN} represent a class of methods that reversely achieve underwater image restoration by learning image degeneration process. In these methods, the optical feature maps, including the attenuation layer, scattering layer, and camera halation, are separately trained to decouple the raw image into corresponding degeneration properties. Besides, they used in-air image as the input to produce pseudo underwater images that are employed as the ground truth to train a supervised restoration network. In contrast, Chen \emph{et al.} \cite{Chen2019} used the atmospheric turbulence model and filtering-based restoration scheme (FRS) to obtain enhanced images and then trained a GAN network in a supervised manner. The above methods are based on supervised learning, so their restoration capacity is largely limited by the pseudo ground truth. 
	
	\subsection{Unsupervised Restoration}
	Researchers have considered learning an underwater image restoration network using style transfer. In particular, restoration of underwater images and degeneration of unpaired in-air images form a closed loop through self-supervision, thus avoiding the need of paired image samples. Li \emph{et al.} \cite{LiCY2018} proposed a restoration transfer method based on weak supervision without the consideration of an optical model. They introduced specific evaluation information, such as contrast and edges, into the loss function, which improved the quality of the restored images. However, when dealing with underwater images with different degeneration effects, color distortion issues are incurred by this physics-free approach. To further improve restoration performance and generalization, we fuse the Jaffe-McGlamery model with the style transfer framework. With underwater RGB images as inputs, we design a generator consisting of multiple sub-networks to produce scene depth map, backscatter factor, attenuation factor, and veiling light. We leverage forward and inverse optical models to build a cyclic training framework. Furthermore, we jointly optimize multiple loss functions and apply a physics-based training paradigm to strengthen the intrinsic constraints of the network.

	\section{Methodology}
	\begin{figure*}
		\centering
		\includegraphics[width=0.85\linewidth]{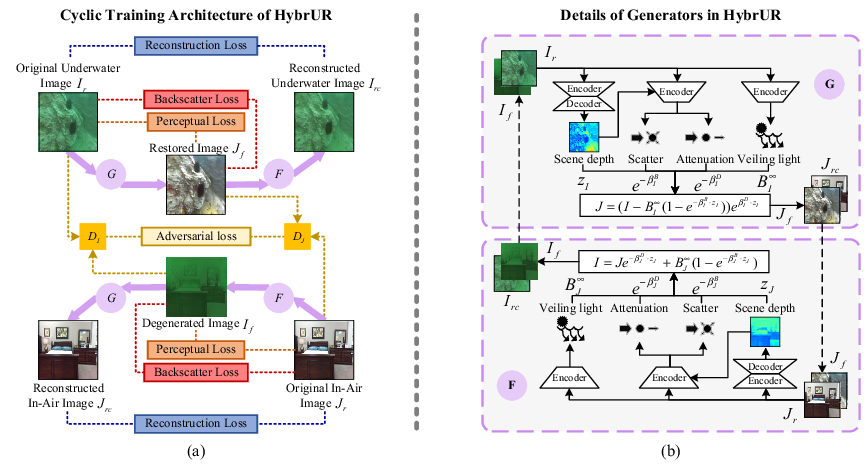}   
		\caption{Architectural schematics and details of generators in HybrUR. (a) In HybrUR, \emph{G} and \emph{F} represent restoration and degeneration models, respectively. The underwater image is restored first and then reconstructed, while the in-air image is degenerated first and then reconstructed. Except for adversarial and reconstruction losses, the backscatter and perceptual losses are introduced. (b) \textit{G} and \textit{F} decompose input images \textit{I} or \textit{J} into attenuation factor, backscatter factor, veiling light, and the scene depth map, and then combine them using the forward-inverse Jaffe-McGlamery model, respectively. With $I_r$ or $I_f$ as input, $G$ produces $J_f$ or $J_{rc}$, respectively. With $J_r$ or $J_f$ as input, $F$ generates $I_f$ or $I_{rc}$, respectively. $G,F$ are trained to improve restoration and degeneration effects utilizing unpaired underwater and in-air images.}
		\label{fig:architecture}
	\end{figure*}
	In this section, a hybrid physical-neural underwater image restoration framework (HybrUR) is discussed. The data flow and implementation of the model are introduced in detail. Finally, the loss function for training is reported.
	\subsection{Overall Architecture}
	
	For brevity of exposition, we first introduce important symbolical notations. There are two generators in our architecture, i.e., restoration network $G$ and degeneration network $F$. In addition, we use $I\in \mathbb{R}^{3\times256\times256}$ and $J\in \mathbb{R}^{3\times256\times256}$ to represent the images with or without underwater degeneration properties, respectively. Also, three subscripts are employed to distinguish the images in data flow, i.e., $I_r,J_r$ indicate real images; $I_f,J_f$ denote fake images generated with real images; $I_{rc},J_{rc}$ are called reconstructed images generated based on fake images. Mathematically, $I_f=F(J_r),J_{rc}=G(I_f),J_f=G(I_r),I_{rc}=F(J_f)$. The data flow is illustrated in Fig.~\ref{fig:architecture}(a).
	As shown, our proposed framework includes two pipelines. In the upper pipeline, the image restoration and reconstruction of underwater images are performed. Correspondingly, the degeneration and reconstruction of in-air images are carried out in the below one. These pipelines utilize a pair of identically structured generative models $G$ and $F$. $G$ is an image restoration generator that can handle both real underwater images and pseudo degenerated images, whereas $F$ represents an image degeneration generator handling in-air or restored underwater images. Therefore, it is expected that the generative effects of \textit{G} and \textit{F} can be enhanced synchronously through the single-stage cyclic training of the unpaired images; \textit{G} finally serves as the underwater image restoration network for inference. Specifically, the quality of restored images can be assessed using two symmetrical discriminators and further improved by jointly optimizing four loss functions. In addition to the conventional adversarial losses, we propose the backscatter loss to improve the backscatter estimation of the degenerated image.
	
	Furthermore, we consider constructing a generator architecture based on underwater image formation model to decouple and control the generation of content and style in image restoration. In general, underwater image formation is generally affected by the optical distortion effect in a nonhomogeneous medium, which can be explicitly given by:
	\begin{align}
		I=A+B,
		\label{eqn:1}
	\end{align}
	where $I$ indicates an underwater RGB image, $A$ is called attenuated image that represents the direct signal which contains the attenuated information, while $B$ represents the scattering mask. Attenuation, known as light absorption, is the main reason for the color distortion of degenerated underwater images. Scattering includes forward and backward effects, which are caused by suspended particles. As reported in \cite{Akkaynak2019}, forward scattering does not contribute significantly to the degeneration of an image, whereas backscattering leads to low global contrast.
	
	Based on the above analysis, we employ the Jaffe-McGlamery model to build the degeneration process with componentized networks, which is expressed as
	\begin{align}
		I=Je^{-\beta^{D}_{J}\cdot z_{J}}+B^{\infty}_{J}(1-e^{-\beta^{B}_{J}\cdot z_{J}}).
		\label{eqn:2}
	\end{align}
	With the input of in-air image $J_r$ or restored underwater image $J_{f}$, the output is the pseudo degenerated image $I_f$ or $I_{rc}$.
	This model is mainly affected by four factors: scene depth $z\in \mathbb{R}^{1\times256\times256}$, veiling light $B^{\infty}\in \mathbb{R}^{3\times1\times1}$, attenuation factor $\beta^{D}\in \mathbb{R}^{3\times1\times1}$, and backscatter factor $\beta^{B}\in \mathbb{R}^{3\times1\times1}$. Note that in some previous studies, it was assumed that $\beta^{D}=\beta^{B}$ to simplify parameter optimization \cite{Wang2017ICIP, Wang2019}. Instead, we consider distinguishing $\beta^{D}$ from $\beta^{B}$ with different sub-networks according to their individual physical definitions \cite{Akkaynak2019, Akkaynak2018}. In addition, both veiling light and degeneration factors maintain different values related to the RGB channels. We further deduce the underwater image restoration model based on Eq.~(\ref{eqn:2}), which is expressed as
	\begin{align}
		J=(I-B^{\infty}_{I}(1-e^{-\beta^{B}_{I}\cdot z_{I}}))e^{\beta^{D}_{I}\cdot z_{I}}.
		\label{eqn:3}
	\end{align}
	
	As shown in Fig.~\ref{fig:architecture}(b), a pair of symmetrically identical flow charts presents two inverse inferences with \emph{G} and \emph{F}. The images at the top left and bottom right indicate the original underwater image $I_r$ and original in-air image $J_r$, respectively. First, the depth maps $z_{I_r}$ and $z_{J_r}$ and the veiling light coefficients $B^{\infty}_{I_r}$ and $B^{\infty}_{J_r}$ are generated by feeding $I_r$ and $J_r$ into the depth networks and veiling light encoders of the two pipelines. Thereafter, the degeneration factor sub-networks produce the attenuation factors $e^{-\beta^{D}_{I_r}},e^{-\beta^{D}_{J_r}}$ and backscatter factors $e^{-\beta^{B}_{I_r}},e^{-\beta^{B}_{J_r}}$ with the concatenated inputs of $\left[I_r,\;z_{I_r}\right]$ or $\left[J_r,\;z_{J_r}\right]$, respectively. The restored image $J_f$ and degenerated image $I_f$ are obtained using Eqs.~(\ref{eqn:3}) and (\ref{eqn:2}), respectively. In addition, we estimate the adversarial and perception losses between the original images and generated results. The dotted lines indicate that pseudo degenerated and restored images are transmitted as inputs into the other generators to achieve image reconstruction, which follows the same process as the previous operation. Finally, we obtain the reconstructed underwater image $I_{rc}$ and reconstructed in-air image $J_{rc}$ and compute their reconstruction losses based on pixel-wise $L_1$ error. 
	
    Furthermore, we propose sub-networks to implicitly associate physical definitions with optical properties. The design and training configurations of the sub-networks are introduced as follows.
		
	\subsection{Scene Depth Map}
	The scene depth network is used for generating a pixel-level range map for a single underwater or in-air image. Depth is essentially defined as the range in the horizontal direction, as described in Eq.~(\ref{eqn:2}); however, we apply it to scenes in different directions assuming small deviations \cite{Akkaynak2019}. Based on the data flow, the depth affects both attenuation and scattering, impacting the content of the generated results. Nevertheless, it is hard and unnecessary to achieve an accurate depth estimation in physical scale because of 2D-to-3D ambiguity in monocular images.	We further argue that the state-of-the-art monocular depth estimation algorithms are not suitable for underwater scenes, since their training data do not involve underwater image domains. Hence, we train a depth sub-network from scratch, which is not to provide an  accurate physical depth but to produce a depth map that fits the degeneration factors for joint optimization.
	
	\begin{figure}
	\centering
	\includegraphics[width=1\linewidth]{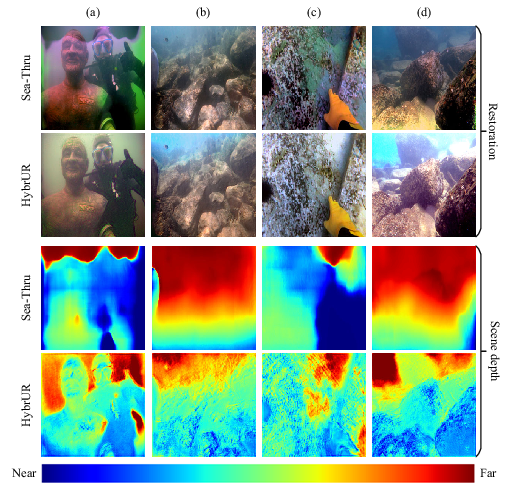}
	\caption{Scene depth estimation and restoration using Sea-Thru and HybrUR. The restored results above show that superior color correction and brightness are conducted by HybrUR. The comparison of depth maps visually explains the color confusion of the Sea-Thru algorithm owing to the limitation of aquatic depth estimation, whereas this issue is addressed in HybrUR.}
	\label{fig:depestimation}
	\end{figure}
	
	In practice, we construct the depth sub-network with two downsampling layers and six residual blocks for deep feature extraction. To avoid the exploding gradient problem caused by exponential growth of the attenuation factor, we regulate the depth value to a range from 0.1~m to 6~m with respect to the training set. In a preliminary experiment, we qualitatively compare the restoration performance and depth estimation of Sea-Thru and HybrUR in Fig.~\ref{fig:depestimation}. As shown, the results of Sea-Thru contain different levels of unacceptable errors in depth and restored color, whereas our method provides more faithful depth information and color restoration. That is, since Sea-Thru employ an off-the-shelf depth estimation network without the training on underwater scenarios, an inaccurate depth projection could be generated for underwater images. In contrast, the proposed depth network in HybrUR can learn more effective depth maps through joint optimization, which is conducive to the color restoration.
	
	\subsection{Veiling Light}
	According to Kannala's modeling \cite{Kannala2006}, veiling light is mainly related to the waveband, spectral reflection, camera properties, and the distance from the imaging location to the water surface. Thus, the output of veiling light sub-network separately models to RGB channels, and we ignore the camera halation since the dataset does not provide camera information. To avoid the overfitting issue, we limit the model size by setting the channel size per convolutional layer to a maximum of 256. In addition, we use the sigmoid activation at the last layer and clamp the output to the range from 0.6 to 1 in the post-processing step.
	
	\subsection{Attenuation and Scattering}
	\label{sec:3-3}
	The attenuation and scattering factors mainly reflect the style characteristics of underwater images. Their explicit expressions are as follows:
	\begin{equation}
		\beta^{D}=\ln\left[\frac{\int_{\lambda_{1}}^{\lambda_{2}}S(\lambda)\rho(\lambda)E(d,~\lambda)e^{-\beta(\lambda)z}d\lambda}
		{\int_{\lambda_1}^{\lambda_2}S(\lambda)\rho(\lambda)E(d,~\lambda)e^{-\beta(\lambda)(z+\Delta z)}d\lambda}\right]/\Delta z,
		\label{eqn:7}
	\end{equation}\vspace{1mm}
	\begin{equation}
		\beta^{B}=-\ln\left[1-\frac{\int_{\lambda_{1}}^{\lambda_{2}}S(\lambda)B^{\infty}(\lambda)(1-e^{-\beta(\lambda)z})d\lambda}
		{\int_{\lambda_1}^{\lambda_2}S(\lambda)B^{\infty}(\lambda)d\lambda}\right]/z.
		\label{eqn:8}
	\end{equation}
	In particular, except for vertical ambient light $E$, the two degeneration factors are both related to the spectral reflection $S$, beam attenuation coefficient $\beta$, waveband $\lambda$, reflectance $\rho$, and depth $z$. The coefficients of $\beta^{D}$ and $\beta^{B}$ have intrinsic correlation and discrepancy. Yet, it is difficult for a single network to simultaneously represent the correlation and discrepancy; therefore, we design partially shared encoding layers to describe the correlation and utilize the independent output layers to learn the discrepancy. As for the network design, we employ the same configuration as that of the veiling light sub-network, except for the input and output layers. Noticeably, considering that both $\beta^{D}$ and $\beta^{B}$ are positive, we employ the sigmoid function as the output layer to produce values from 0 to 1 as the terms $e^{-\beta^{D}}$ and $e^{-\beta^{B}}$.
	Since the parameter size of the degeneration factor sub-networks are fewer than that of the depth network, the convergence of degeneration factors is faster than that of depth map, leading to poor performance or even mode collapse of the depth map.
	To solve this problem, we feed a concatenated input $[I,\;z_{I}]$ or $\left[J,\;z_{J}\right]$ into the degeneration factor sub-networks to improve the training stability and facilitate the convergence of depth map. 
	
	\subsection{Loss and Training} \label{sec:3-4}
	The loss function for training our framework is as follows:
	\begin{equation}
		L=\lambda_{g}L_{GAN}+\lambda_{c}L_{cycle}+\lambda_{p}L_{perc}+\lambda_{B}L_{\hat{B}},
		\label{eqn:10}
	\end{equation}
	where $L_{GAN}$ and $L_{cycle}$ indicate the adversarial loss and reconstruction loss, respectively. $L_{perc}$ represents the perceptual loss to compensate for edge imperfections. $L_{\hat{B}}$ is used to improve the stability during training. Moreover, $\lambda_g$ $\lambda_c$, $\lambda_p$, and $\lambda_{B}$ are used to balance the effects of multiple loss functions.
	
	We employ the classic CycleGAN loss to perform the self-supervision mechanism of a cyclic architecture, which is expressed as
	\begin{equation}
		\begin{split}
			L_{cycle}(G,F)=&\mathbb{E}_{I_r\sim p_{data}(I_r)}\left[\left\|F\left(G\left(I_r\right)\right)-I_r\right\|_{1}\right]\\
			&+\mathbb{E}_{J_r\sim p_{data}(J_r)}\left[\left\|G\left(F\left(J_r\right)\right)-J_r\right\|_{1}\right],
			\label{eqn:11}
		\end{split}
	\end{equation}
	where $I_r$ and $J_r$ indicate the original underwater and in-air images, respectively. Mao \emph{et al.} \cite{Mao2017} suggested that the loss function based on cross entropy might lead to the vanishing gradient and utilized the minimum mean square error to train the GAN framework, which is given by
	\begin{equation}
		\begin{split}
			L_{D_I}=&\mathbb{E}_{I_r\sim p_{data}(I_r)}\left[(D_I(I_r)-1)^{2}\right]\\
			&+\mathbb{E}_{J_r\sim p_{data}(J_r)}\left[D_I(F(J_r))^{2}\right],
			\label{eqn:12}
		\end{split}
	\end{equation}\vspace{.1mm}
	\begin{equation}
		L_{F}=\mathbb{E}_{J_r\sim p_{data}(J_r)}\left[(D_I(F(J_r))-1)^{2}\right],
		\label{eqn:13}
	\end{equation}
	where $L_{D_I}$ and $L_F$ denote the discrimination error and generation error for images with degeneration properties. Accordingly, the loss functions of discriminator $D_J$ and generator $G$ can be constructed in similar forms. The adversarial loss in cyclic generation framework is as follows:
	\begin{equation}
		L_{GAN} = L_G + L_F+ L_{D_I} + L_{D_J}.
		\label{eqn:13-add-14}
	\end{equation}
	
	In addition, $D_I,D_J$ are discriminators, built with the Conv-BatchNormalization-ReLU (CBR) modules. Inspired by PatchGAN, we construct the discrimination branch with the idea of \emph{Patch}\cite{Isola2017}, where the patch scale is related to the number of CBR modules. In practice, we use three CBR modules to build the discriminator, resulting in the patch scale of 1/3 original resolution.
	
	However, generating ideal restored images by using these typical losses is not enough. Since the $L_1$ loss function is sensitive to the depth imperfections and discordance of degeneration factors, we add a perceptual loss to solve this problem. We employ an off-the-shelf image encoder $e$ (i.e., VGG16) to extract the representation $e^{k}(I)\in \mathbb{R}^{C_k\times W_k\times H_k}$, where $\Omega_k=\left\{0,...,W_k-1\right\}\times\left\{0,...,H_k-1\right\}$ is the corresponding spatial domain, and $\left|\;\cdot\;\right|$ counts the pixel numbers of the spatial domain \cite{LiFF2016}. The perceptual loss is formulated as follows:
	\begin{equation}
		L^{(k)}_{perc}=-\frac{1}{\left|\Omega_k\right|}\sum_{u,v\in\Omega_k}\left(e^{k}_{u,v}(s_r)-e^{k}_{u,v}(s_f)\right)^{2}, s\in \left\{I,\;J\right\}.
		\label{eqn:14}
	\end{equation}
	We calculate the mean square error for each pixel index $u,v$ in the $k$-th layer. In practice, we have verified that individually applying $ReLU_{3-3}$ from the VGG16 to calculate the loss could preserve adequate edge details of the restored image.
	
	To further regulate the generated scene depth, veiling light, and backscatter factor that conform to the image formation properties \cite{CVPR2004Schechner}, we propose a backscatter loss function. Previous literature show quantitatively that backscattering increases exponentially with depth and eventually reaches saturation in underwater scenes \cite{Akkaynak2018}. When all light is absorbed, completely shadowed, or far from the camera, the captured RGB intensity $I_{R, G, B}$ approaches the backscatter mask $B_{R, G, B}$. Therefore, we first divide the pixels of the original image into ten groups $\Gamma_{i}=\left\{\left[d^i, d^{i+1}\right],\;i=0,1,2,\cdots,9\right\}$ in order of depth $z$ and then create a mask $M$. $M_{u,r}=1$ at a coordinate set $\left\{(u,r)\right\}$ that contains the lowest 1\% pixels of average RGB density $\left(I^{R}_{u,r}+I^{G}_{u,r}+I^{B}_{u,r}\right)/3$ in each depth group $\Gamma_{i}$. These selected pixels are applied to estimate the backscattering and calculate the difference from the values obtained using Eq.~(\ref{eqn:2}), which is described as
	\begin{equation}
		\hat{B}_{I}\odot M=B^{\infty}_{I}(1-e^{-\beta^{B}_{I}\cdot z_{I}})\odot M,
		\label{eqn:9}
	\end{equation}
	where $\hat{B}_{I}$ represents the estimated backscattering, while $\odot$ denotes the batchwise Hadamard product. In unsupervised training, this constraint provides an additional loss function. The scene depth, veiling light, and backscatter factor can be further optimized by decreasing the $L_1$ loss of the left and right terms in Eq.~(\ref{eqn:9}). Therefore, the backscatter loss function is given by
	\begin{equation}
		L_{\hat{B}}=\mathbb{E}\left[\left\|I\odot M - B^{\infty}_{I}\left(1-e^{-\beta^{B}_{I}\cdot z_{I}}\right)\odot M\right\|_{1}\right].
		\label{eqn:15}
	\end{equation}	
	
	\section{Experiments and Discussion}
	In this section, we first present training details. To verify the superiority of the physics- and data-driven restoration architecture, we conduct a comprehensive ablation study by removing one component at a time. The visualization comparison of backscatter estimation and depth estimation illustrates the modeling and interpretability of the generator, confirming the advantages of the hybrid physical-neural framework. A comparison of the restored results on three public datasets, U45, RUIE, and UIEB, demonstrates that the proposed method presents better results for a variety of underwater image quality metrics. In particular, 
	our results are more visually faithful than compared methods. In addition, we verify the enhancement performance through several underwater vision-based application tests, including key point extraction and matching.
	
	\subsection{Training Details}
	We adopt the same RUIE training set as in \cite{LiCY2018}. Considering preliminary validation and testing on RUIE, we randomly select 2076 images for training, while the remaining images serve as validation and test samples. Different from the in-air images used in \cite{LiCY2018}, we sample the same number of images from the indoor vision dataset Sun RGB-D. All images are used directly without data augmentation. A batch size of 16 is used. For better generalization, we construct the input with random underwater image and in-air image.
	
	Since the proposed generator consists of four sub-networks, the training parameters are set differently from those of CycleGAN. For the depth sub-network, degeneration factor sub-networks, and discriminators, the learning rates are 0.0002, 0.0001, and 0.0001, respectively, which linearly decay from the 30th epoch. The Adam optimizer is initialized with $\beta=\{0.5, 0.999\}$. Note that the specific weights of different loss functions are crucial since they strongly affect the convergence and restoration effect. In general, $L_G$ is responsible for the visual effect of the generative results, whereas $L_{cycle}$, $L_{perc}$, and $L_{\hat{B}}$ primarily guarantee stability during training. If $L_{cycle}$, $L_{perc}$, and $L_{\hat{B}}$ are too large, the depth network converges quickly, and a poor depth map is produced, resulting in mode collapse. To alleviate the color deviation to a large extent, we emphasize the generative error and relatively weaken the cyclic consistency and perception errors. $\lambda_g=3$, $\lambda_c=4$, $\lambda_p=0.1$, and $\lambda_B=2$ are identified experimentally based on the model performance.
		
	\subsection{Comparison Methods}
	We compare the proposed method with some prior-art methods. Three dehaze-based traditional image enhancement methods are discussed in comparison experiments, namely DCP \cite{DCP2010}, UDCP \cite{UDCP2013}, and WCID \cite{WCID2011}. Regarding learning-based methods, four model-based supervised approaches are compared: FUnIE-GAN \cite{Islam2020}, UWGAN-UIE \cite{Wang2019}, GAN-RS \cite{Chen2019}, and the deep image formation model (DIFM) \cite{Chen2021}. In addition, we further compare our method with an unsupervised method Water-Net \cite{LiCY2018} to demonstrate the improvement in unsupervised training introduced by our approach. All compared methods are implemented using open-source codes. We separately compare them on the U45, UIEB, which consist of various water types, and the RUIE dataset, which contains turbid underwater images. In particular, for UWGAN-UIE, we only select the model that best fits the underwater images in our training set from their pretrained networks.
	\begin{table}[!t]
		\renewcommand\arraystretch{1.25}
		\newcommand{\tabincell}[2]{\begin{tabular}{@{}#1@{}}#2\end{tabular}}
		\scriptsize
		\centering
		\caption{\label{tab:1}Ablation study}		
		\begin{tabular}{l|l|c|cccc}
			\Xhline{1.2pt}
			Dataset & Method    & Sharpness $\uparrow$  & $\sigma_c$ $\uparrow$ & $con_l$ $\uparrow$ & $\mu_s$ $\uparrow$ & UCIQE $\uparrow$ \\ \hline
			&\scriptsize Original & 0.363 & 0.017 & 0.476 & 0.797 & 0.344 \\
			& CycleGAN & 0.528 & 0.024 & 0.902 & 0.771 & 0.457  \\ \cline{2-7}
			& w/o $z$ & 0.564 & 0.029 & 0.694 & \textbf{0.802} & 0.411 \\
			& w/o \tiny$B^{\infty}$ & 0.614 & 0.041 & 0.880 & 0.796 & 0.466 \\
			& w/o \tiny$e^{-\beta^{D}}$& 0.588 & 0.032 & 0.839 & 0.783 & 0.447 \\
			RUIE&w/o \tiny$e^{-\beta^{B}}$ & 0.605 & 0.038 & 0.855 & 0.789 & 0.456 \\
			& \scriptsize w/o \tabincell{l}{\tiny Loss$_{\hat{B}}$,\\ \tiny Joint input} & 0.607 & 0.044    & 0.892  & 0.790  & 0.469 \\
			&\scriptsize  w/o \tiny Loss$_{\hat{B}}$  & 0.624 & 0.053    & 0.898  & 0.793  & 0.476 \\
			&  \scriptsize w/o \tiny Joint input    & 0.619 & 0.055    & 0.912  & 0.790  & 0.480 \\
			& \scriptsize Ours full   & \textbf{0.629} & \textbf{0.061}    & \textbf{0.966}  & 0.780  & \textbf{0.495} \\\hline
				
			&\scriptsize Original & 0.516 & 0.043 & 0.569 & 0.779 & 0.377\\
			& CycleGAN & 0.581 & 0.031 & 0.843 & 0.711 & 0.429  \\ \cline{2-7}
			& w/o $z$ & 0.650 & 0.056 & 0.737 & \textbf{0.794} & 0.433 \\
			& w/o $B^{\infty}$ & 0.713 & 0.075 & 0.865 & 0.777 & 0.473 \\
			& w/o $e^{-\beta^{D}}$& 0.671 & 0.061 & 0.836 & 0.753 & 0.452 \\
			U45&w/o $e^{-\beta^{B}}$ & 0.698 & 0.071 & 0.848 & 0.755 & 0.460 \\
			&\scriptsize w/o \tabincell{l}{\tiny Loss$_{\hat{B}}$, \\ \tiny Joint input} & 0.704 & 0.077    & 0.877  & 0.731  & 0.465 \\
			& \scriptsize w/o \tiny Loss$_{\hat{B}}$  & 0.736 & 0.091    & 0.882  & 0.781  & 0.486 \\
			& \scriptsize w/o \tiny Joint input & 0.729 & 0.080    & 0.914  & 0.730  & 0.476 \\
			& \scriptsize Ours full   & \textbf{0.744} & \textbf{0.099}    & \textbf{0.917}  & 0.771  & \textbf{0.497} \\ \Xhline{1.2pt}
		\end{tabular}
	\end{table}
	
	\subsection{Ablation Study} \label{sec:4-3}
	\begin{figure}[!t]
		\centering
		\includegraphics[width=0.9\linewidth]{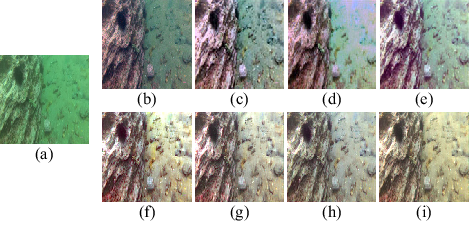}
		\caption{Examples of the ablation experiment. (a) Original image. (b)--(i) Images correspond to the ablated models listed below CycleGAN in Table~\ref{tab:1}.}
		\label{fig:3}
	\end{figure}
	For the lack of ground-truth restored images, some no-reference quality assessment metrics are employed in the ablation study, including sharpness and underwater color image quality evaluation (UCIQE) \cite{Yang2015}. In addition, considering the influence of absorption and scattering masks on image brightness, contrast, and sharpness, two metrics for assessing contrast and sharpness employed in \cite{JVCIR2018} are used to justify the visual quality enhancement of our model. The UCIQE is formulated via the following equation:
	\begin{equation}
		UCIQE = c_1 * \sigma_c + c_2 * con_l + c_3 * \mu_s,
		\label{eqn:16}
	\end{equation}
	where $\sigma_c$, $con_l$, and $\mu_s$ denote standard deviation of the chroma, brightness contrast, and mean value of saturation, respectively. $c_1=0.468$, $c_2=0.2745$, and $c_3=0.2576$ are corresponding weights given in \cite{Yang2015}.
	
	To understand the influence of the individual components in HybrUR, we remove one component at a time and evaluate the performance of the ablated model on two different test sets, as shown in Table~\ref{tab:1}. Visual examples of each ablated term in the table are arranged sequentially in Fig.~\ref{fig:3}.
	Table~\ref{tab:1} presents the ablation results for two test sets with different image types. First, $\sigma_c$ and $\mu_s$ of the baseline decrease in both datasets compared to the original data, indicating that it is difficult to use single end-to-end unsupervised training to effectively learn the color correction model. Owing to the lack of a specific architecture or loss function to retain image content, CycleGAN fails to reconstruct the image structure while learning the style transformation. In contrast, better results are obtained for all ablated models compared with those of the original images. For rows 3--5 in each dataset, we employ all four loss functions and the concatenated input of the degeneration factor sub-networks. Thereafter, the four network components are replaced individually with constant counterparts of the same size. ``w/o $z$'' indicates that we map the uniformly distributed noise to the depth range to replace the predicted depth map. Despite the use of noisy depth map, HybrUR can be regularly trained in this situation. However, the foreground and background are mixed up, and poor color correction is observed when the constant depth map is used, resulting in low metrics for both sharpness and UCIQE. Additionally, although the best saturation for both datasets is obtained using ``w/o $z$'', we do not consider it as a key assessment for color correction because excessive saturation is not positively correlated to a good visual result, as shown in Fig.~\ref{fig:3}(b). ``w/o $B^{\infty}$'', ``w/o $e^{-\beta^{D}}$'', and ``w/o $e^{-\beta^{B}}$'' denote that we utilize the constant parameters [0.5, 0.5, 0.5], respectively, in place of the corresponding network outputs. As shown in row ``w/o $e^{-\beta^{D}}$'', the restoration process is substantially affected by attenuation due to its exponential product term, resulting in partial pixels exceeding the maximum gray level, thus destroying the detailed textures and color distribution. In contrast, the veiling light and backscatter sub-networks have relatively less influence on the restoration in terms of UCIQE. However, as can be seen from $con_l$ that the backscatter factor sub-network can effectively eliminate the backscattering mask, resulting in improved image contrast. As shown in rows 6--8, the validity of the backscatter loss, concatenated in
	put of the degeneration factor sub-networks, and the complete architecture of HybrUR are evaluated. Backscatter loss is not utilized in ``w/o Loss$_{\hat{B}}$''. As shown, the correlation between the depth, veiling light and backscatter factor sub-networks is not well-learned, resulting in color distortion. For ``w/o Joint input'', the attenuation factor and backscatter factor sub-networks are trained with RGB images. As a result, the improvements of UCIQE brought by concatenated input and the backscatter loss are remarkable in multiple datasets. In terms of sharpness enhancement, the concatenated input results in better performance than the backscatter loss. Overall, based on the visual comparison shown in Fig.~\ref{fig:3}, the Jaffe-McGlamery model-based hybrid physical-neural architecture and physics-based training paradigm are both crucial components for the improvement of comprehensive restoration performance, where the hybrid physical-neural architecture affects more than the physics-based training paradigm of HybrUR.
	
	\subsection{Correlation Between Depth and Degeneration Factors}
	\begin{figure}
		\subfigure[]{
			\begin{minipage}[c]{0.41\linewidth}
				\centering
				\includegraphics[width=1\linewidth]{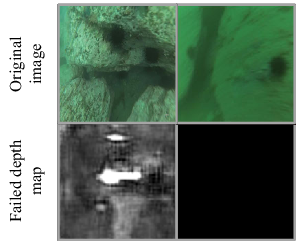}
			\end{minipage}
			\label{fig:4a}
		}	
		\subfigure[]{
			\begin{minipage}[c]{0.5\linewidth}
				\centering
				\includegraphics[width=1\linewidth]{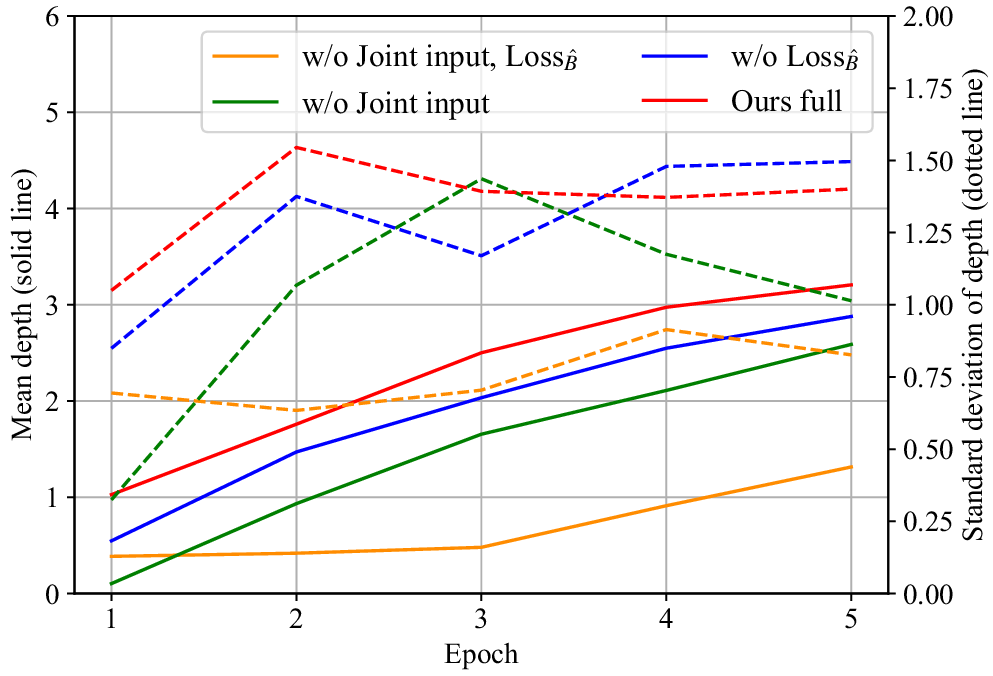}
			\end{minipage}
			\label{fig:4b}
		}\quad
		\subfigure[]{
			\begin{minipage}[c]{1\linewidth}
				\centering
				\includegraphics[width=0.95\linewidth]{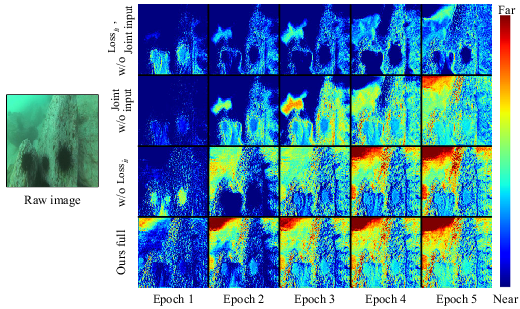}
			\end{minipage}
			\label{fig:4c}
		}	
		\caption{Mode collapse and qualitative analysis of generated depth map. (a) Mode collapse types of generated depth map. (b) Mean depth and standard deviation of test sets in the first five epochs. (c) Generated depth map sequences in the first five epochs.}
		\label{fig:4}
	\end{figure}
	Fig.~\ref{fig:4a} illustrates two types of depth map collapse that appear in unsupervised restoration. The left column indicates the obstinate patches in the depth map. Assuming that the degeneration factors and depth map are trained independently by updating the degeneration factors during early training, fake images can easily deceive the discriminators. Hence, the depth network cannot be sufficiently trained. In extreme cases, the depth map may converge to an unacceptable black map, as shown in the right column. According to Eqs.~(\ref{eqn:2}) and (\ref{eqn:3}), since $e^{-\beta^{D}\cdot z},e^{-\beta^{B}\cdot z}$ equal to 1 with $z=0$, the degeneration factors became useless in inference and training.
	\begin{figure}
		\centering
		\includegraphics[width=0.9\linewidth]{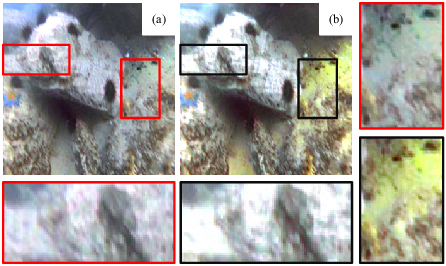}
		\caption{Qualitative comparison of concatenated input for degeneration factor sub-networks. (a) An image restored by HybrUR trained with concatenated input. (b) An image restored by HybrUR trained without concatenated input. The selected boxes highlight reductions of discordance between depth and degeneration factors by using the concatenated input.}
		\label{fig:5}
	\end{figure}
	
	To verify the effectiveness of the concatenated input in avoiding depth map collapse, we compare four training configurations of ``w/o Loss$_{\hat{B}}$, Joint input'', ``w/o Loss$_{\hat{B}}$'', ``w/o Joint input'', and ``Ours full'', and visualize the depth maps of a random underwater image for the first five training epochs. As shown in Fig.~\ref{fig:4c}, the foreground and background of the depth maps are incorrectly produced when using ``w/o Loss$_{\hat{B}}$, Joint input'', which eventually converge to incorrect patches. As shown in rows 2 and 3, both Loss$_{\hat{B}}$ and concatenated input positively influenced the production of reasonable gradients. In contrast, a higher improvement in depth estimation is observed for ``w/o Loss$_{\hat{B}}$'' than ``w/o Joint input''; hence, the concatenated input effectively improved the learning of the depth network during early training stage. In addition, we calculate the mean depth and standard deviation of the test sets during the first five epochs, as shown in Fig.~\ref{fig:4b}. The solid line represents the mean depth values, and the dashed line denotes the standard deviation of the depth map. The results of the concatenated input show that the statistical mean of the depth values is closer to the median of 3m, and the depth values are more evenly distributed throughout the map. Furthermore, we qualitatively compare the restoration results with and without concatenated input. Fig.~\ref{fig:5}(a) shows the restored image sampled from `Ours full', and Fig.~\ref{fig:5}(b) shows that sampled from `w/o Joint input'. We use red and black boxes to highlight two typical areas. The comparison results at the bottom emphasize some over-exposures caused by the discordance of the depth and degeneration factors, and the right boxes show that the use of concatenated input can avoid unexpected color distortion.
	
	\subsection{Backscatter Loss}
	\begin{figure}
		\centering
		\includegraphics[width=1\linewidth]{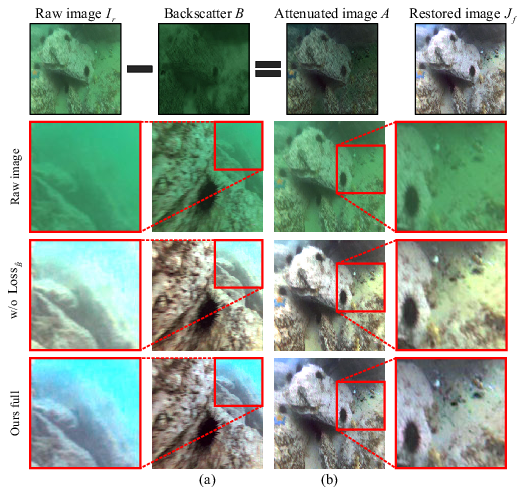}
		\caption{Visualization of backscattering elimination and qualitative comparison of the image restoration effects trained with and without backscatter loss.}
		\label{fig:6}
	\end{figure}
	We present quantitative and qualitative comparisons for Loss$_{\hat{B}}$ to verify the enhancement in the restoration performance. The results of the qualitative comparison are shown in Fig.~\ref{fig:6}. First, by substituting the generated depth map, veiling light, and backscatter factors into $\hat{B_{I}} = B^{\infty}_{I}(1-e^{-\beta^{B}_{I}\cdot z_{I}})$, a dark green backscattering mask $B$ is obtained, the hue of which is consistent with the ambient color of the original image. The comparison of the restoration results show a brighter visual effect with ``w/o Loss$_{\hat{B}}$''; however, the actual performance is poor based on the quantitative metrics. The details are highlighted in the red boxes. For ``Ours full'', with the fading of the dim shadow at the distance, the edges of the reef and seabed stand out through the dissociation of the backscattering mask from the original image. In contrast, the details of `w/o Loss$_{\hat{B}}$' do not appear owing to the over-exposure in the distant scene, which is caused by the discordance of depth and degeneration factors.
	
	A numerical comparison of the visual examples is provided in Table~\ref{tab:2}. We compare the contrast and sharpness of both the attenuated and restored images. In particular, the contrast and sharpness of the attenuated image $A$ are substantially increased compared with the original image $I$, indicating that elimination of backscatter masks can effectively improve the quality of the underwater image. A larger increase in the result obtained by ``Ours full'' indicates that the backscatter loss function can facilitate the estimated backscatter mask to closely mimic real optical scattering effects, thus improving the details of hazy areas and preserving color fidelity well.
	
	\subsection{Comparison of Restoration Quality}
\begin{figure*}
		\centering
		\includegraphics[width=1\linewidth]{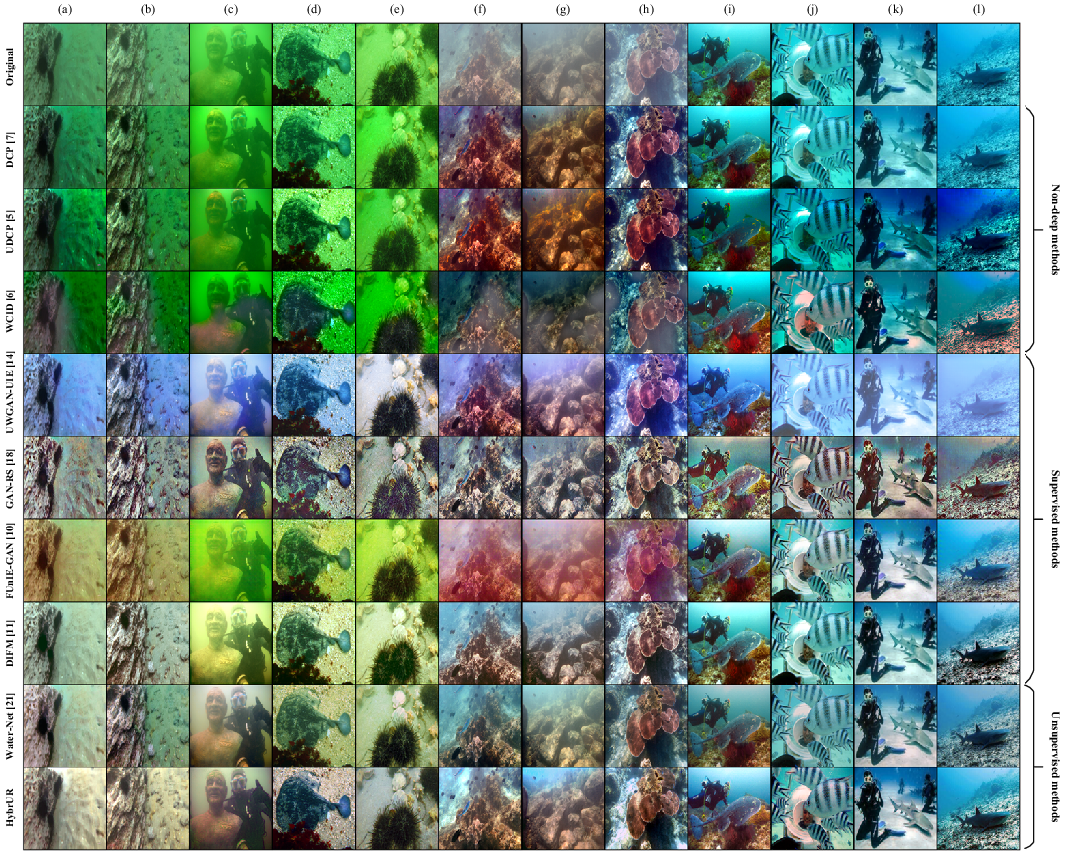}  
		\caption{Qualitative comparison between HybrUR and contemporary approaches in terms of restoration quality. Each row is the generated results using the indicated method. The top three rows under originals are the results of the non-deep methods. The succeeding four rows represent the supervised learning, and the last two rows are unsupervised methods. The twelve columns show the underwater images with different degeneration styles, where the images in the first two columns are from RUIE, those in columns (c)--(g) are from U45, and those in columns (h)--(l) are from UIEB.}
		\label{fig:7}
	\end{figure*}

	To demonstrate the robustness of the proposed method, the restoration results for the green, hazy and blue underwater images are illustrated in Fig.~\ref{fig:7} (from left to right). Visually, the proposed HybrUR exhibit distinct advantages in terms of balanced color distribution and stretched contrast. We evaluate the non-deep, supervised learning, and unsupervised learning methods separately. The traditional methods considered in the comparison are all associated with the dehaze model; therefore, the haze is effectively eliminated with these methods, as shown in columns (f), (g), and (h). However, the color distortion problems is not addressed. In contrast, the supervised underwater image restoration methods performed well in several certain degeneration types, yet the generalization to various image domains is limited. For example, DIFM is not suitable for green-style image restoration, as shown in columns (c)--(e). FUnIE-GAN cannot solve the hazy problem caused by backscattering, as shown in columns (f)--(h). Proper color correction cannot be achieved for blue-distorted images with GAN-RS. As an unsupervised restoration network, Water-Net has a uniform and comprehensive restoration capability, except for deeply green-distorted images. Compared with these results, our method achieves a more realistic color correction effect. In addition, the blurred degeneration in columns (f)--(i) is not well-addressed with Water-Net. In contrast, HybrUR can significantly improve the contrast and color brightness of the restored images by estimating and removing the backscatter mask.
	
\begin{table}[!t]
		\renewcommand\arraystretch{1.25}
		\footnotesize
		\centering
		\caption{\label{tab:2}Quantitative comparison of image contrast and sharpness}
		\begin{tabular}{m{7mm}<{\centering}|l|c|c}
			\Xhline{1.2pt}
			Label                & Terms   & Contrast $\uparrow$ & Sharpness $\uparrow$ \\ \hline
			\multirow{5}{*}{(a)} & $I$       & 0.166   & 0.503             \\ \cline{2-4}
			& $A_{\text{w/o~Loss}_{\hat{B}}}$    & 0.192   & 0.5             \\
			& $A_{\text{Ours full}}$ & \textbf{0.214}   & \textbf{0.532}             \\ \cline{2-4}
			& $J_{\text{w/o~Loss}_{\hat{B}}}$    & 0.246   & 0.777            \\
			& $J_{\text{Ours full}}$ & \textbf{0.251}   & \textbf{0.799}            \\ \hline
			\multirow{5}{*}{(b)} & $I$       & 0.142   & 0.486            \\ \cline{2-4}
			& $A_{\text{w/o~Loss}_{\hat{B}}}$    & 0.182   & 0.512           \\
			& $A_{\text{Ours full}}$ & \textbf{0.218}   & \textbf{0.557}            \\ \cline{2-4}
			& $J_{\text{w/o~Loss}_{\hat{B}}}$    & 0.235   & 0.761            \\
			& $J_{\text{Ours full}}$ & \textbf{0.260}  & \textbf{0.793}            \\ \Xhline{1.2pt}
		\end{tabular}
		\begin{tablenotes}
			\footnotesize
			\item Note: (a) and (b) correspond to the two image samples in Fig.~\ref{fig:6}.
		\end{tablenotes}
	\end{table}

	\begin{table*}[!t]
		\renewcommand\arraystretch{1.25}
		\small
		\centering
		\caption{\label{tab:3}Quantitative comparison using no-reference quality assessment}		
		\begin{tabular}{l|l|cccc|cccc|cc}
			\Xhline{1.2pt}
			Dataset               & Method  & $\sqrt{d_o}$$\downarrow$    & $d_ad_b$$\uparrow$      & $a_l$     & $U$$\downarrow$              & $\sigma_c$$\uparrow$       & $con_l$$\uparrow$         & $\mu_s$$\uparrow$ & UCIQE$\uparrow$   & Contrast$\uparrow$ & Sharpness$\uparrow$      \\ \hline
			& Original   & 0.386 & 0.059          & 53.192 & 0.200          & 0.043          & 0.569          & 0.779  & 0.377        &0.196 &0.516  \\
			&DCP \cite{DCP2010} &0.468 &0.136 &43.062 &0.1 &0.058  &0.711 &0.822 &0.434 &0.285 &0.605 \\
			&UDCP \cite{UDCP2013} &0.42 &0.223 &33.042 &0.072 &0.069 &0.785 &\textbf{0.883} &0.475 &\textbf{0.392} &0.635 \\
			&WCID \cite{WCID2011} &0.353 &0.224 &38.236 &0.054 &0.082 &0.618 &0.864 &0.431 &0.32 &0.602 \\ \cdashline{2-12}
			\multirow{2}{*}{U45}  & UWGAN-UIE \cite{Wang2019}& 0.393 & 0.077          & \textbf{56.141} & 0.110          & 0.049          & 0.606          & 0.756  & 0.384   &0.177 &0.589       \\
			&FUnIE-GAN \cite{Islam2020} &0.444 &0.134 &48.181 &0.08 &0.059 &0.648 &0.814 &0.415 &0.231 &0.642\\
			& GAN-RS \cite{Chen2019}  & \textbf{0.129} & 0.138          & 49.285 & \textbf{0.020}          & 0.057          & 0.777          & 0.818  & 0.450  &0.241 &\textbf{0.939}        \\
			& DIFM \cite{Chen2021}     & 0.288 & 0.083          & 53.402 & 0.084          & 0.065          & 0.838          & 0.790  & 0.464    &0.277 &0.75      \\\cdashline{2-12}
			& Water-Net \cite{LiCY2018} & 0.252 & 0.082          & 46.188 & 0.074          & 0.064          & 0.777          & 0.821  & 0.455        &0.269 &0.663  \\
			& HybrUR    & 0.364 & \textbf{0.255} & 54.239 & 0.032          & \textbf{0.099} & \textbf{0.917} & 0.771  & \textbf{0.497} &0.329 &0.744 \\ \hline
			& Original   & 0.431 & 0.016          & 51.273 & 0.639          & 0.017          & 0.476          & 0.797  & 0.344    &0.142 &0.363      \\
			&DCP &0.437 &0.081 &38.041 &0.16 &0.031 &0.669 &0.874 &0.423 &0.3 &0.488\\
			&UDCP &0.396 &0.146 &28.111 &0.121 &0.036 &0.653 &\textbf{0.918} &0.432 &\textbf{0.332} &0.452\\
			&WCID &0.473 &0.171 &40.257 &0.082 &0.032 &0.444 &0.886 &0.365 &0.262 &0.415\\ \cdashline{2-12}
			\multirow{2}{*}{RUIE} & UWGAN-UIE & 0.430 & 0.040          & 49.037 & 0.235          & 0.041          & 0.657          & 0.780  & 0.401      &0.204 &0.543    \\
			&FUnIE-GAN &0.424 &0.073 &50.937 &0.134 &0.036 &0.654 &0.818 &0.407 &0.187 &0.583\\
			& GAN-RS   & \textbf{0.124} & 0.087          & 50.766 & 0.031          & 0.041          & 0.806          & 0.810  & 0.449    &0.225 &\textbf{0.919}      \\
			& DIFM      & 0.239 & 0.027          & 55.884 & 0.191          & 0.035          & 0.780          & 0.782  & 0.432   &0.225 &0.607       \\\cdashline{2-12}
			& Water-Net & 0.200 & 0.045          & 48.846 & 0.119          & 0.047          & 0.810          & 0.818  & 0.455      &0.242 &0.596    \\
			& HybrUR    & 0.205 & \textbf{0.149} & \textbf{57.955} & \textbf{0.028} & \textbf{0.061} & \textbf{0.966} & 0.780  & \textbf{0.495}&0.262 &0.629 \\\hline
			& Original  & 0.408 & 0.097     & 45.3 & 0.152          & 0.05          & 0.634          & 0.797  & 0.403          &0.224 &0.565\\
			&DCP &0.388 &0.123 &41.244 &0.119 &0.061 &0.731 &0.832 &0.443 &0.297 &0.622\\
			&UDCP &0.333 &0.215 &30.9 &0.078 &0.071 &0.81 &\textbf{0.898} &0.487 &\textbf{0.438} &0.641\\
			&WCID &0.318 &0.188 &37.8 &0.061 &0.086 &0.67 &0.862 &0.446 &0.332 &0.633\\ \cdashline{2-12}
			\multirow{2}{*}{UIEB} & UWGAN-UIE & 0.556 & 0.112     & 52.849  & 0.115      & 0.046          & 0.602          & 0.74  & 0.378         &0.159 &0.591 \\
			&FUnIE-GAN &0.349 &0.136 &46.035 &0.072 &0.065 &0.714 &0.242 &0.439 &0.256 &0.673 \\
			& GAN-RS   &\textbf{0.151} &0.15       & 49.041  &\textbf{0.022}          & 0.056          & 0.79          & 0.818  & 0.454    &0.238 &\textbf{0.915}      \\
			& DIFM     &0.288 & 0.1      &47.8  & 0.084     & 0.074        &\textbf{0.858}      & 0.803  &\textbf{0.477}   &0.292 &0.741   \\\cdashline{2-12}
			& Water-Net & 0.281 & 0.095    &45.18 & 0.081   & 0.062       & 0.745      & 0.844  & 0.451  &0.278 &0.677     \\
			& HybrUR   & 0.186   & \textbf{0.233} & \textbf{52.953} &0.035 & \textbf{0.089} &0.811 & 0.825  &\textbf{0.477} &0.408 &0.758  \\  \Xhline{1.2pt}
		\end{tabular}
	\end{table*}
	
	Furthermore, a numerical comparison is presented in Table~\ref{tab:3}. Some no-reference metrics are employed, including contrast, sharpness, UCIQE, and an underwater index in the LAB color space. This index is proposed by Chen \emph{et al.} to evaluate the color distribution in the LAB color space and is expressed as follows \cite{Chen2019}:
	\begin{align}
		\label{eqn:17}
		U=\frac{\sqrt{d_{o}}}{a_{l}d_{a}d_{b}},
	\end{align}
	where $a_l$ is the mean value of the light channel; $d_o$ is the distance between the center of the image color distribution and the LAB space origin; $d_a$ and $d_b$ are the image color spans along the $a$ and $b$ axes, respectively. Consequently, $U$ can reflect the degree of the concentration and bias for a color distribution. Table~\ref{tab:3} lists three test sets: U45, RUIE, and UIEB. We use dotted lines to distinguish between traditional, supervised, and unsupervised methods. The data consistently confirm the results of the qualitative analysis discussed above. Dehaze-based traditional methods effectively increase the contrast of the restored image; however, color deviation remains an issue to be solved. For the supervised methods, DIFM has the best UCIQE value for UIEB but has weak scores for RUIE, owing to the lack of green image labels in the training set. Since its multistage training and specific losses for image enhancement, GAN-RS achieves the highest sharpness in each test, which is better than its color restoration effect. Aiming at unsupervised restoration, Water-Net has low $con_l$ and contrast, because the backscatter mask is not fully eliminated (see Fig.~\ref{fig:7}(c) and (g)). In contrast, our model has the widest color span in the LAB space, indicating its natural visual effect. Also, our method exhibits superior color correction and illumination variance compared to the other approaches, thus achieving the best UCIQE for all test sets. In addition, we further evaluate these approaches using a UIQM metric\cite{UIQM}, and the average results for the three datasets are shown in Table.~\ref{tab:uiqm}, where our model obtains comparable results compared to Water-Net. In particular, HybrUR has notably good results in terms of color correction, whereas Water-Net performed better for edge generation. The average UIQM results of the supervised methods are generally higher than those of unsupervised frameworks due to their advantages in terms of UISM. In summary, HybrUR has difficulty in retaining edges of underwater images; however, it accurately estimates and eliminates backscatter, thereby improving the color correction and holistic sharpness.
	\begin{table}[!t]
		\renewcommand\arraystretch{1.25}
		\newcommand{\tabincell}[2]{\begin{tabular}{@{}#1@{}}#2\end{tabular}}
		\small
		\centering
		\caption{\label{tab:uiqm}Quantitative comparison using UIQM assessment}		
		\begin{tabular}{l|cccc}
			\Xhline{1.2pt}
			Method   & UICM & UISM & UIConM  & UIQM \\ \hline
			\makecell[l]{Supervised methods \\(average among\\\cite{Chen2019, Chen2021, Wang2019, Islam2020 })} & 4.508 & 7.136 & 0.282 & 3.243\\ \cline{1-5}
			Water-Net \cite{LiCY2018} & 5.142 & 6.709 & 0.286 & 3.149  \\
			HybrUR & 7.583 & 6.413 & 0.283 & 3.119 \\
			\Xhline{1.2pt}
		\end{tabular}
	\end{table}
	
	\subsection{Feature-Extraction Tests}
\begin{figure}
		\centering
		\includegraphics[width=0.88\linewidth]{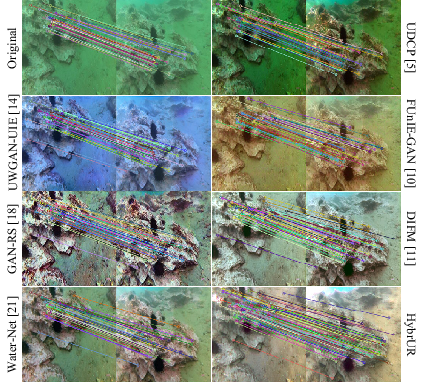}  
		\caption{Comparison of SIFT key points matching. The results shown correspond to methods in Fig.~\ref{fig:7}.}
		\label{fig:8}
	\end{figure}

	In this subsection, we focus on the representation performance of underwater image enhancement and restoration algorithms. For a more comprehensive evaluation, we apply several algorithms to evaluate their generated results from the perspective of low-level features, including the scale-invariant feature transform (SIFT) key point and Harris corner \cite{SIFT}. Fig.~\ref{fig:8} shows a visual example of the SIFT matching on raw image pair and restored results.
	As shown, the restoration result of HybrUR is better, providing more matched key point pairs than Water-Net and some supervised methods. The average numbers of key points and Harris corners detected by the above methods are compared in Table~\ref{tab:4}. GAN-RS performs the best on both datasets due to its highest sharpness, while HybrUR performs better in the evaluation of unsupervised methods. Image enhancement using HybrUR can still be improved, which will be the focus of our future work.
	
	\subsection{Discussion}	
	Underwater image quality restoration is regarded as an essential technology for vision-based underwater localization and navigation. We propose a novel unsupervised training framework based on the Jaffe-McGlamery model to achieve color correction and image quality improvement in degenerated underwater images. We verify the effectiveness of the hybrid physical-neural framework and demonstrate that the physics-based training paradigm, including concatenated input of the degeneration factor sub-network and a backscatter loss function, is conducive to stable training and restoration performance via a thorough ablation experiment. Both quantitative and qualitative comparisons prove the superiority of HybrUR. All models in HybrUR are implemented with PyTorch, whose training and testing is based on a GeForce GTX 1080ti GPU. The GPU speed of HybrUR reaches 33.17 frames per second (FPS), which is much higher than that of Sea-Thru (0.16~FPS) that is also based on the Jaffe-McGlamery model and Water-Net (5.88~FPS). To further enhance restoration performance of the hybrid physical-neural framework, two aspects are considered: 1) the unsupervised generation network generally has poor performance in restoring edge details; thus, a specific loss functions or super-resolution methods could be designed to improve the pixel-level accuracy; 2) the network training is not effective enough on a dataset with extremely diverse distribution of underwater domains. This can be improved by referring to a many-to-many style transfer method \cite{Lee2018}.
	\begin{table}[!t]
		\renewcommand\arraystretch{1.25}
		\small
		\centering
		\caption{\label{tab:4}Average numbers of SIFT key points and Harris corners}	
		\begin{tabular}{l|l|c|c}
			\Xhline{1.2pt}
			Dataset               & Method    & SIFT & Harris \\ \hline
			\multirow{8}{*}{RUIE} & Original    & 182  & 222    \\
			& UDCP \cite{UDCP2013}     & 358  & 232    \\\cdashline{2-4}
			& UWGAN-UIE \cite{Wang2019} & 306  & 245    \\
			& FUnIE-GAN \cite{Islam2020} & 473  & 256    \\
			& GAN-RS \cite{Chen2019}    & 883  & 367    \\
			& DIFM \cite{Chen2021}      & 664  & 283    \\\cdashline{2-4}
			& Water-Net \cite{LiCY2018} & 443  & 247    \\
			& HybrUR    & 608  & 251    \\ \hline
			\multirow{8}{*}{U45}  & Original    & 34   & 247    \\
			& UDCP      & 138  & 250       \\\cdashline{2-4}
			& FUnIE-GAN & 244  & 260       \\
			& UWGAN-UIE & 200  & 256    \\
			& GAN-RS    & 425  & 375    \\
			& DIFM      & 352  & 267    \\ \cdashline{2-4}
			& Water-Net & 259  & 254    \\
			& HybrUR     & 356  & 262    \\ \Xhline{1.2pt}
		\end{tabular}
	\end{table}

	\section{Conclusions and Future Work}
	In this paper, we have creatively proposed an unsupervised method that can learn an image restoration model for degenerated underwater images from unpaired underwater and in-air image datasets. This method is constructed using the CycleGAN architecture and explicit physical inferences based on the Jaffe-McGlamery degeneration model. Degenerated underwater images are decoupled into content and style, which include scene depth, backscatter factor, attenuation factor, and veiling light. These individual elements are reversely composed to generate a color-corrected image, resulting in better restoration performance than a single end-to-end generator. We have demonstrated the effectiveness of the proposed hybrid physical-neural solution and proved that the concatenated input of the degeneration factor networks and backscatter loss are strong cues for the enhancement of the depth estimation, global contrast, and color correction, which synthetically contribute to restoration. The comparison experiments demonstrate that our method outperforms the state-of-the-art unsupervised underwater image restoration network and some supervised approaches. In addition, the low-level feature applications demonstrate the practice value of the proposed method.
	
	In future work, we will primarily focus on enhancing image edge details in unsupervised color correction. Furthermore, we will construct a large-scale original underwater image dataset with underwater robots and investigate a style transfer from multiple input and output domains.

	\vspace{-10mm}
	\begin{IEEEbiography}[{\includegraphics[width=1in,height=1.25in,clip,keepaspectratio]{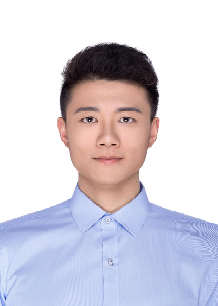}}]{Shuaizheng Yan}
		received the B.E. degree in measurement technology and instruments from University of Electronic Science and Technology of China, Chengdu, China, in 2018, and the Ph.D. degree in control theory and control engineering from the Institute of Automation, Chinese Academy of Sciences (IACAS), Beijing, China, in 2023.
		
		He is currently an Associate Professor with the Department of Mechanical Engineering, Fuzhou University. His research interests include underwater perception, intelligent control, and navigation systems of underwater robots.
	\end{IEEEbiography}
	\vspace{-13mm}
	\begin{IEEEbiography}[{\includegraphics[width=1in,height=1.25in,clip,keepaspectratio]{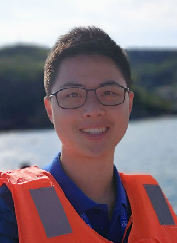}}]{Xingyu Chen}
		received the B.E. degree in electrical engineering and automation from the College of
		Nuclear Technology and Automation Engineering, Chengdu University of Technology, Chengdu, China, in 2015, and the Ph.D. degree in control theory and control engineering from the Institute of Automation, Chinese Academy of Sciences (IACAS), Beijing, China, in 2020.
		
		He is currently a Researcher with Xiaobing.AI. His research interests lie in the joint field of robotics, graphics, and computer vision, including but not limited to human digitalization, 3D geometry understanding, and human-machine interaction.
	\end{IEEEbiography}
	\vspace{-13mm}
	\begin{IEEEbiography}[{\includegraphics[width=1in,height=1.25in,clip,keepaspectratio]{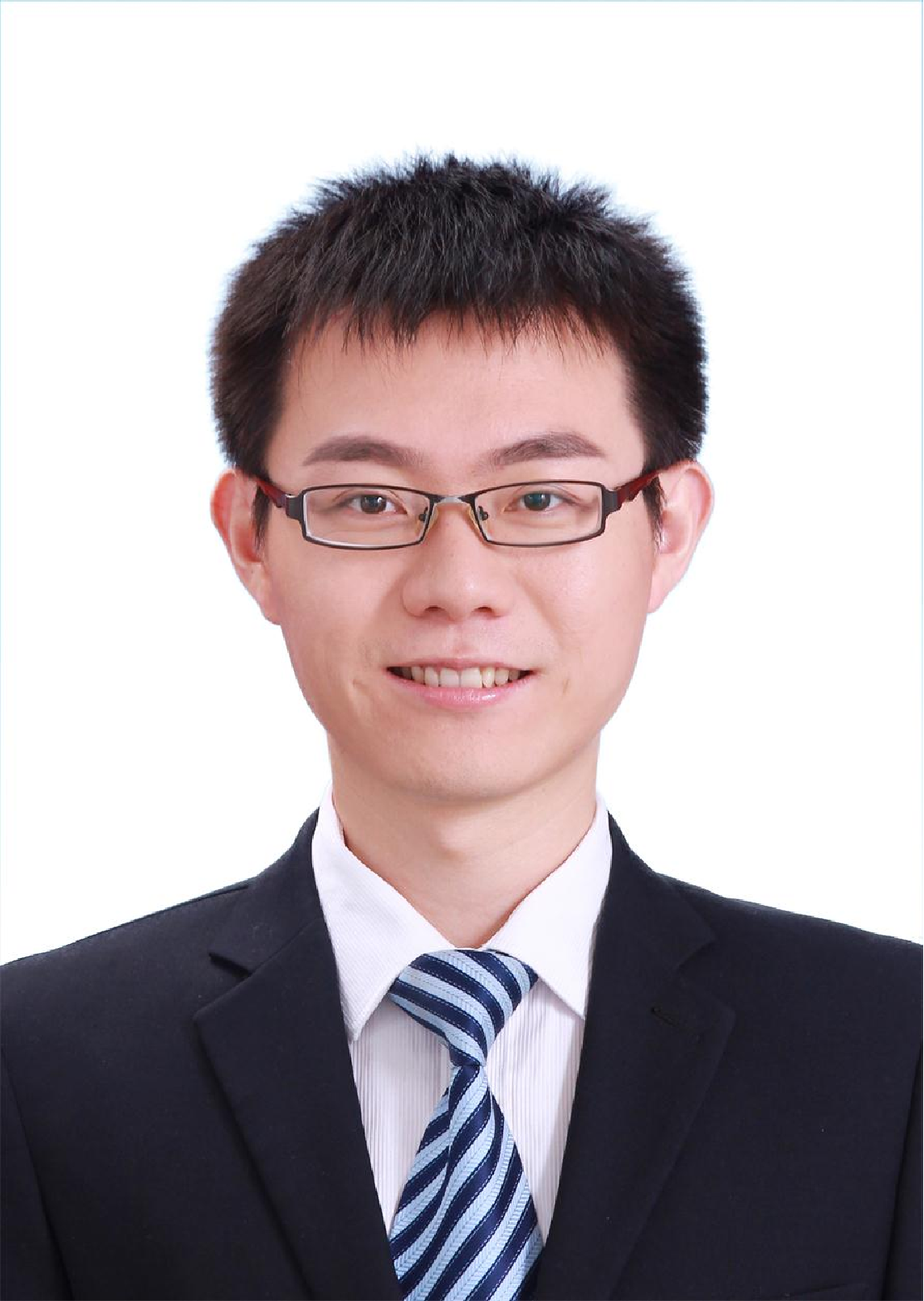}}]{Zhengxing Wu}
		received the B.E. degree in logistics engineering from the School of Control Science and Engineering, Shandong University, Jinan, China, in 2008, and the Ph.D. degree in control theory and control engineering from the Institute of Automation, Chinese Academy of Sciences (IACAS), Beijing, China, in 2015.
		
		He is currently a Professor with the Laboratory of Cognitive and Decision Intelligence for Complex System, IACAS. His current research interests include bioinspired robots and intelligent control systems.
	\end{IEEEbiography}
	\vspace{-12mm}
	\begin{IEEEbiography}[{\includegraphics[width=1in,height=1.25in,clip,keepaspectratio]{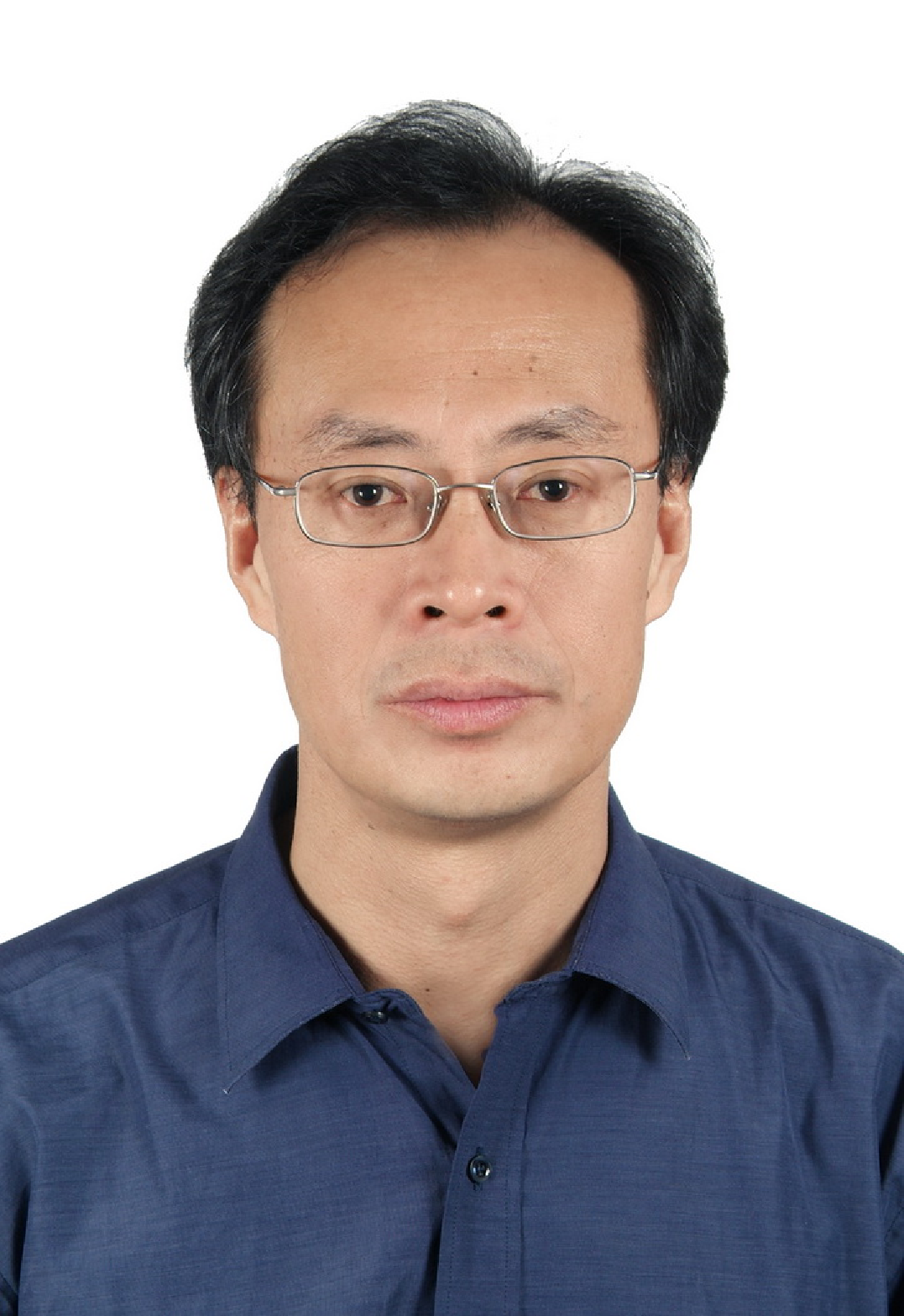}}]{Min Tan}
		received the B.Sc. degree from Tsinghua University, Beijing, China, in 1986, and the Ph.D. degree from the Institute of Automation, Chinese Academy of Sciences (IACAS), Beijing, China, in 1990, both in control science and engineering. He is currently a Professor with the Laboratory of Cognitive and Decision Intelligence for Complex System, IACAS. He has published more than 200 papers in journals, books, and conference proceedings.
		
		His research interests include robotics and intelligent control systems.
	\end{IEEEbiography}
	\vspace{-10mm}
	\begin{IEEEbiography}[{\includegraphics[width=1in,height=1.25in,clip,keepaspectratio]{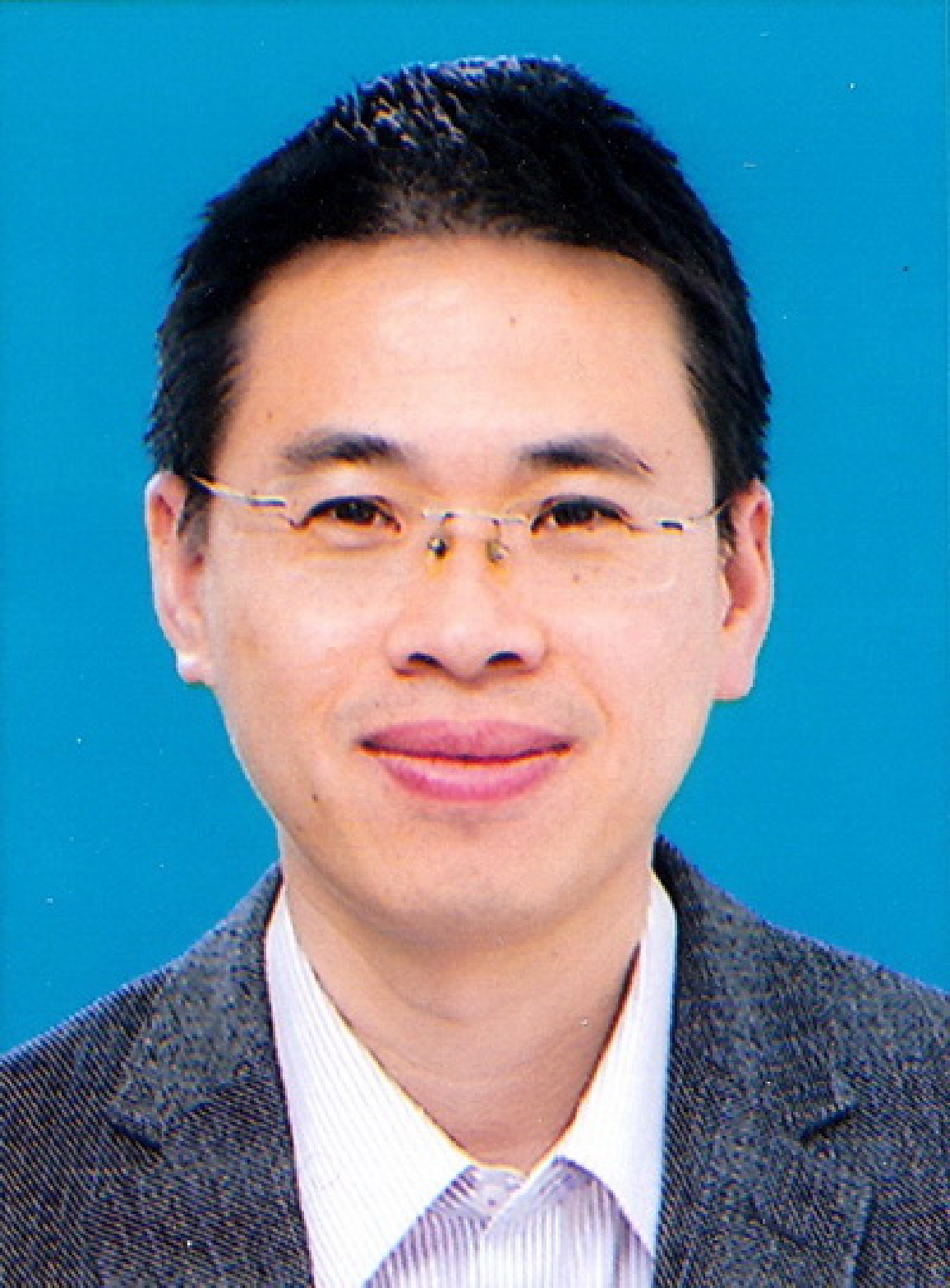}}]{Junzhi Yu}
		(IEEE Fellow) received the B.E. degree in safety engineering and the M.E. degree in precision instruments and mechanology from the North University of China, Taiyuan, China, in 1998 and 2001, respectively, and the Ph.D. degree in control theory and control engineering from the Institute of Automation, Chinese Academy of Sciences, Beijing, China, in 2003.
		
		From 2004 to 2006, he was a Post-Doctoral Research Fellow with the Center for Systems and Control, Peking University, Beijing. He was an Associate Professor with the Institute of Automation, Chinese Academy of Sciences, in 2006, where he was a Full Professor in 2012. In 2018, he joined the College of Engineering, Peking University, as a Tenured Full Professor. His current research interests include intelligent robots, motion control, and intelligent mechatronic systems.
	\end{IEEEbiography}
\end{document}